\documentclass[10pt,letterpaper]{article}

\usepackage{cogsci}

\cogscifinalcopy 

\usepackage[backend=biber,style=apa,annotation=false,natbib=true]{biblatex}
\usepackage{booktabs}
\usepackage{bbm}
\usepackage{graphicx}
\usepackage[table]{xcolor}
\usepackage{tabularx}

\usepackage{comment}

\title{Similarity All The Way Up: Multilingual Generalization in LLMs Relies on Language-Level Similarity Structures}

\author{
  {\large\bfseries Supantho Rakshit$^1$, Adele Goldberg$^2$, Henry Conklin$^3$} \\
  {\normalsize\normalfont
     $^1$Department of Electrical and Computer Engineering, 
     $^2$Department of Psychology,
     $^3$Laboratory for Artificial Intelligence\\
     Princeton University
   }
 }

\begin{document}

\maketitle

\begin{abstract}
As Large Language Models (LLMs) grow more capable across diverse tasks, their (in)ability to generalize remains difficult to quantify and poorly understood beyond limited domains. In particular, LLMs are known to struggle generalizing multilingually, to languages outside of English, and that are poorly attested in their training data. To understand why this may be, and what enables some models to perform better than others, we turn to a long history of work across the cognitive sciences, arguing that successful generalization derives from appropriate representations in similarity space. We look at how well LLMs' representations capture the hierarchical similarity structure between distinct languages. Strikingly, we show LLMs' latent representations largely recover the hierarchical structure of the Indo-European language family tree -- grouping languages that are members of the same subfamily closely together in representation space. Furthermore, we show that the degree to which models reflect the similarity structure of languages correlates with their performance on XNLI, a multilingual natural language inference benchmark. This extends classic work on similarity-driven generalization at scale, showing how models that \textit{represent similar languages similarly} generalize better from one language to another.

\end{abstract}

\section{Introduction}

Large Language Models (LLMs) perform impressively across a broad range of tasks, but we still lack a clear understanding of how their representational structures drive their performance. Multilinguality represents an interesting test case. The data used to train contemporary models remains disproportionately in English, and model performance quickly degrades as you move outside high-resource languages (languages with greater than 1 billion tokens of available data, e.g., German, French, Chinese). 
To understand why this may be and what representational structures enable better multilingual performance, we analyze the similarity structure learned by LLMs. We show models implicitly recover similarity relationships among natural languages in their representation space, without explicit supervision. Moreover, we show that this representational structure relates to performance, with models that represent similar languages similarly performing better on a standard benchmark of multilingual performance. 

Generalization, or the ability to make inferences about new cases on the basis of prior experience, has long been understood in cognitive science as dependent on the underlying structure of representations. \citet{shepard1957stimulus, shepard1987toward} first observed that generalization follows an exponential gradient in psychometric space: the probability of relating a novel item to an existing one decays as a function of their distance in representation space.  This principle, later given a rational Bayesian justification by \citet{tenenbaum2001generalization}, shows how 
generalization behavior is determined by representational geometry. Exemplar models of categorization formalized this relationship, showing that generalization to novel items depends on their similarity to stored instances in a multidimensional space \citep{medin1978context, nosofsky1986attention, kruschke1992alcove}, while prototype effects demonstrated that representations encode central tendencies that support interpolation to never-seen items \citep{posner1968genesis}. These findings motivated a more general theoretical claim: that representations simply \emph{are} similarity structures and that cognitive generalization reduces to operations over the geometry of these spaces \citep{edelman1998representation, gardenfors2000conceptual, kriegeskorte2013representational}.

Large Language Models, based on the transformer architecture \citep{vaswani_attention_2017}, are connectionist models, with distributed representations acquired through gradient-based learning. A long history of work has shown how such representations place similar items in nearby regions of activation space, enabling networks to generalize on the basis of learned similarity structure \citep{hinton1986learning, rumelhart1986learning, elman1990finding, rogers2004semantic}. Classic demonstrations showed that networks generalize morphological patterns to novel words \citep{rumelhart1986past} and semantic properties to novel concepts based on the similarity structure of their internal representations.

\begin{figure*}[t]
    \centering
    \includegraphics[width=1\linewidth]{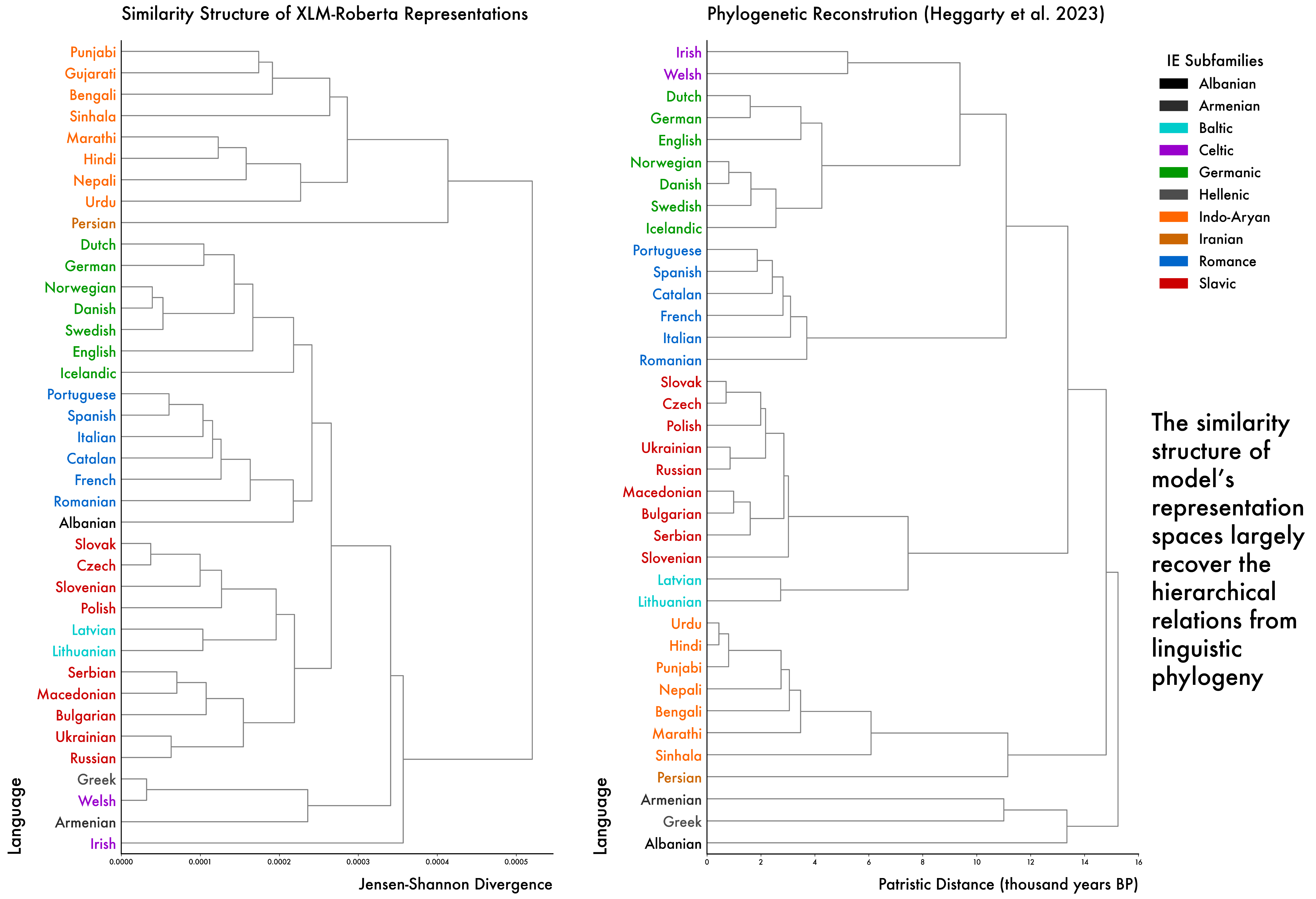}
    \caption{\textbf{LLMs Implicitly Recover Relations Between Languages:} Similarity structure among 38 Indo-European languages, as derived from XLM-Roberta's representations (Left) and reconstructed from \citet{heggarty2023language} (Right). Colors represent phylogenetic language subfamilies, with a legend in the top right.}
    \label{fig:coretree}
\end{figure*}

An existing body of work has studied multilingual LLMs and the extent to which their representations enable cross-lingual transfer. Multilingual LLMs are increasingly trained without parallel data, simply performing next token prediction on data from each language independently \citep{devlin_bert_2019, conneau2020unsupervised}. Such models can often generalise between languages -- when fine-tuned on a specific task in one language, performance improves on the same task when evaluated in other languages as well  \citep{pires2019multilingual}. These performance gains have been found to be greater between languages with shared vocabulary and typological characteristics \citep{wu2019beto}. Having finite representational capacity forces models to learn structures that are effectively shared across languages, with models converging to representations that are not entirely specific to a single language, instead aligning related semantic concepts cross-linguistically \citep{conneau2020unsupervised}. Previous work using mBERT representations of multilingual word lists finds that phylogenetic distance is the strongest predictor of pair-wise representational distances for $100$ languages \citep{rama2020probing}. Here, instead of looking at the pairwise case, we look at whether LLMs recover the full set of similarity structures across languages, exemplified in our case by the Indo-European family tree. Our analysis systematically compares twelve LLMs with varying architectures, aggregating representations over a full set of documents in each language rather than focusing on concept-level word embeddings. 
Further, we predict that models that represent \textit{similar languages similarly} will generalise better across languages.  If the similarity structure of a representation space defines its generalization behaviour \citep{tenenbaum2001generalization}, then representations that reflect hierarchical relations among languages, beyond mutually interpretable languages like German and Dutch, should generalise best. That is, models that group linguistic subfamilies (e.g. Germanic, Slavic, Romance) closely together should perform better at tasks that require cross-linguistic transfer. 

To characterise similarity among languages, we turn to lexico-statistical work on linguistic phylogeny \citep[e.g.][]{bouckaert2012mapping, gray2003language, heggarty2023language}. For instance, \citet{gray2003language} aims to reconstruct the phylogenetic tree of the world's languages from corpora of lexical overlaps.  While statistical approaches have been criticized as being unable to distinguish borrowing from shared ancestry \citep[e.g.][]{nichols2008tutorial,  kassian2025language}, our  goal is to estimate hierarchical similarity relationships among languages rather than mirror the historical evolution of Indo-European.   To this end, it is widely appreciated that contact between languages, as well as shared ancestry, has a strong influence on convergences among languages \citep{haspelmath2005world} and dialects within a language \citep{dunn2025language, wieling2017exploring}. For example, Norwegian and Icelandic are historically sister languages, since both descended from Old West Norse, yet Norwegian does not share Icelandic's case system, complex grammar, nor its writing system, and the two languages are mutually unintelligible. Danish evolved from a distinct dialect (Old East Norse), but is far more similar to Norwegian today than Icelandic is, and written forms of Danish and Norwegian are highly mutually intelligible. Statistical models based on estimated distances between languages provide an independent way of estimating similarity relations among individual languages and  sub-families\citep[e.g.][]{heggarty2023language, bouckaert2012mapping}. Therefore, rather than mapping to a historically veridical tree structure based on the comparative method of \citep{nordhoff2011glottolog}, we adopt the tree structure induced from a statistical model, namely (\citet{heggarty2023language}), as an independent reference tree that 
captures hierarchical similarity relationships among languages.

Here we show that LLMs implicitly recover the structure of the reference tree to a remarkable extent. Further, we show models explicitly trained to be multilingual approximate the reference tree more closely. The current experiments extend a cognitively motivated account of generalization at scale, with similarity structure playing a central role not just for specific objects or words, but across entire languages.

\section{Methods}

Our analysis builds a hierarchical similarity tree based on representations of 38 languages, for each of 12 LLMs. We compare the trees based on each model's representations to the independent reference tree by computing three different distances: subtree and path kernels and cophenetic correlations. We find across models that their representational similarity structure largely recovers the structure of the reference tree. We visualize the structure of one model, XLM-Roberta, in Figure 1 (Left) alongside the reference tree constructed from \citet{heggarty2023language} (Figure 1: Right).

\paragraph{Language sample}
The current work investigates similarity relationships among 38  written, living Indo-European (IE) languages from 5 subfamilies: Romance (6), Germanic (6), Indo-Aryan (9), Slavic (9), Baltic (2), and independent IE languages ([3]: Greek, Armenian, and Albanian). We restrict our attention to IE here because similarities among any pair of IE languages are expected to be non-zero; multiple languages within several language families are included to determine whether models capture smaller degrees of similarity within the higher range accurately.  

\paragraph{The Reference Tree}
To assess the validity of similarity relationships among languages, we use a tree structure constructed on the basis of \citet{heggarty2023language} as the target reference. \citet{heggarty2023language} fit a Bayesian model to lexical cognate judgments for 170 concepts across 161 Indo-European languages. To construct the reference tree for our sample of 38 languages, we extracted the corresponding terminal nodes from the full Bayesian Maximum Clade Credibility (MCC) tree from Heggarty et al. (2023), using exact string matching against the CLDF language metadata. In cases where modern standard varieties were unavailable, we selected the closest dialectal proxy (e.g., WelshNorth for Welsh, ArmenianEastern for Armenian). We computed pairwise patristic distances between all language pairs as: $ d_{\text {patristic }}(i, j)=\sum_{e \in \text { path }(i, j)} \ell(e) $, where path $(i, j)$ denotes the unique path connecting terminals $i$ and $j$ through their most recent common ancestor, and $\ell(e)$ is the branch length (in thousands of years before present) of edge $e$. This yields a symmetric $38 \times 38$ distance matrix $\mathbf{D}_{\text {ref }}$ where entry $D_{i j}$ represents the estimated time depth to the most recent common ancestor of languages $i$ and $j$. We performed UPGMA clustering on $\mathbf{D}_{\text {ref }}$ to produce a dendrogram for visualization and metric computation, maintaining the ultrametric property of the Bayesian phylogeny. 

\paragraph{Measuring LLMs' ability to generalize across languages: XNLI}
We use the Cross-lingual Natural Language Inference (XNLI) benchmark \citep{conneau-etal-2018-xnli}  to quantify the extent to which each model enables zero-shot transfer across languages. XNLI requires models to determine whether a premise sentence entails, contradicts, or is neutral with respect to a hypothesis sentence. Though NLI has been criticized as a somewhat unnatural task \citep{weissweiler2025linguistic}, XNLI is a practical and well-established way to measure how well models generalize from one language to another.

\paragraph{LLM Density Estimation} 
To capture similarity relationships in each model, we estimate a conditional density in the model's embedding space $P(\hat{Z}|\text{language id})$ which describes how a given language is represented. To do so, we employ the soft-density estimator from \citet{conklin2025information}. This approach estimates distributions in continuous space by softly-discretizing them. For a model with $L$ transformer layers and hidden dimension $d$, let $h_t^{(\ell)} \in \mathbb{R}^d$ denote the hidden state for token $t$ at layer $\ell$. We construct a discrete codebook $\mathcal{C} = \{c_1, \ldots, c_K\}$ by sampling uniformly at random from the surface of the unit hypersphere. 
We set $K = 3d$ to ensure consistent expressive capacity across different model architectures. Each token embedding $h_t^{(\ell)}$ is softly assigned to codebook entries via temperature-scaled softmax, then aggregated across all tokens in a language to produce a language-specific conditional distribution. For a given language with token set $\mathcal{T}$, the distribution at layer $\ell$ is:

\begin{equation}
P^{(\ell)}(\hat{Z} \mid \text{language}) = \frac{1}{|\mathcal{T}|} \sum_{t \in \mathcal{T}} \text{softmax}\left(\frac{h_t^{(\ell)} \cdot \mathcal{C}}{\tau \|h_t^{(\ell)}\|_2}\right)
\end{equation}

\noindent where $\tau = 0.1$ is the temperature parameter and the softmax operates over all $K$ codebook entries in $\mathcal{C}$. Hence, $P^{(\ell)}(\hat{Z})$ is an estimate of a categorical distribution-- this is related to kernel density estimation \citep{parzen1962estimation}. By conditioning these estimates $P^{(\ell)}(\hat{Z}| X=x)$, where $x=\text{the language}$ of the corresponding input sentence, we get a robust estimation of how a given language is represented in the model's embedding space at layer $\ell$.

\begin{figure*}[t]
    \centering \includegraphics[width=1\linewidth]{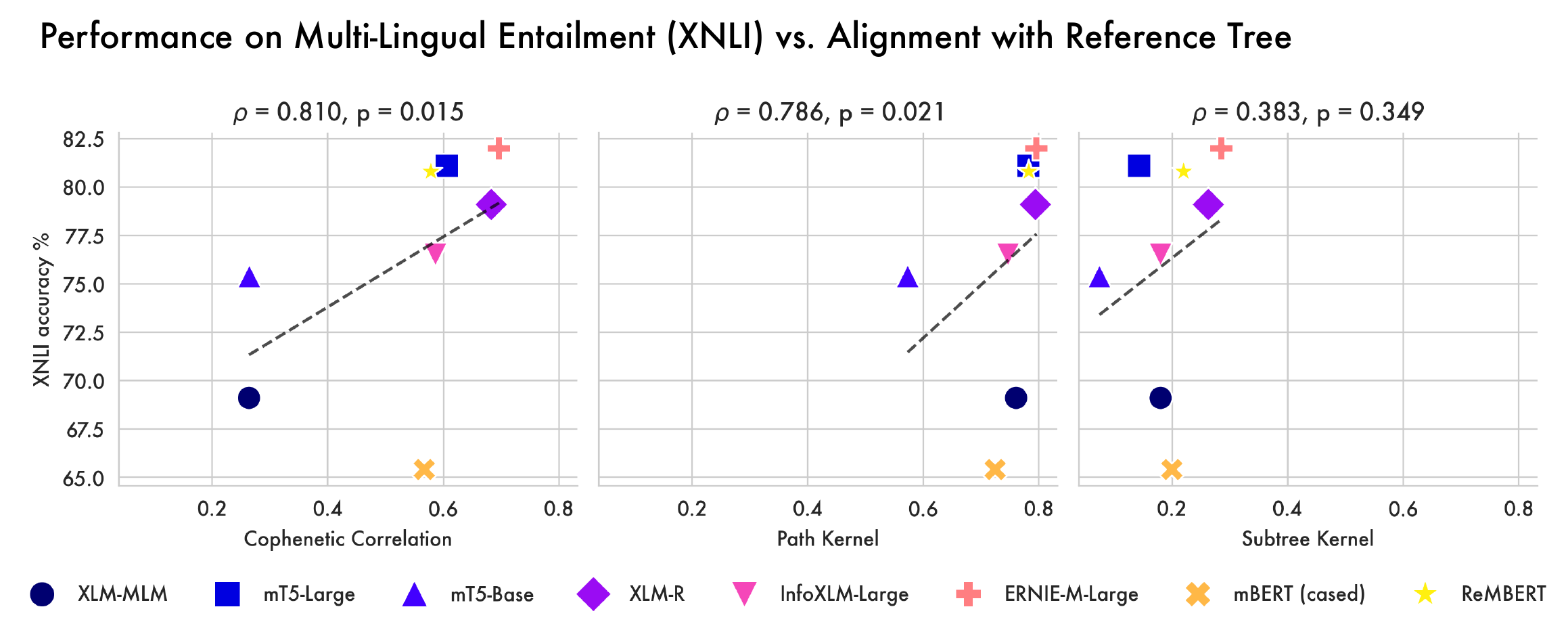}
    \caption{\textbf{Correlating Structure with Performance:} 
    Each facet shows one of three tree-distance measures (\textit{x}-axis) plotted against XNLI accuracy percentage (\textit{y-}axis) for eight LLMs. Above each facet is the Spearman rank correlation $\rho$ between axes.  Cophenetic Correlation and Path Kernel show a significant positive relationship with performance; both reflect how well LLM representations capture higher-order structures of the reference, like sub-family membership, rather than exact neighbor match.}
    \label{fig:xnli}
\end{figure*}

\begin{figure*}[t]
    \centering
    \includegraphics[width=0.95\linewidth]{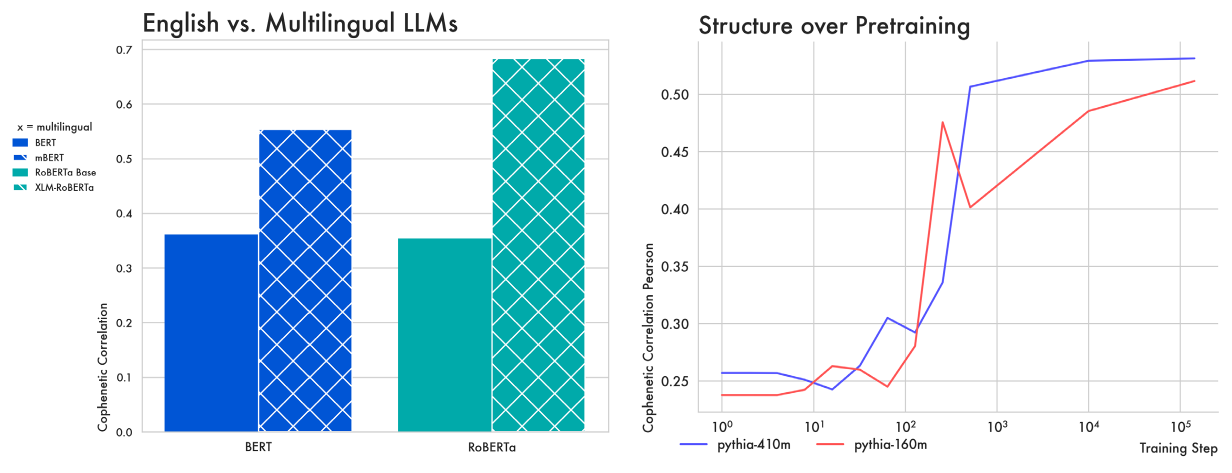}
    \caption{\textbf{Multilingual vs. English LLMs | Structure Over Pretraining:} \textbf{(Left)} \textit{y}-axis: degree of alignment with the reference tree measured by the cophenetic correlation; \textit{x}-axis: one of the model families BERT or RoBERTa. Colors correspond to model IDs. All models share the same architecture, number of layers, and parameters, but differ in training data. Models trained on multilingual data have a white x pattern. Multilingual models better align with the reference tree. \textbf{(Right):} y-axis: Cophenetic Correlation with the reference tree; x-axis: time-points in pretraining an LLM in terms of gradient step (log scaled); Color: size of the Pythia model. We observe that representational structures begin to correlate with the reference tree early on.}
    \label{fig:pretraining}
\end{figure*}

\paragraph{Constructing a Representational Tree} To quantify differences between language distributions, we compute pairwise Jensen-Shannon divergences. Given data in a set of languages $Q$ for each layer $\ell$ and language pair $(q_i, q_j)$, we compute their layer-wise divergence, aggregating across layers to get $d$:


\begin{equation}
d(q_i, q_j) = \frac{1}{L}\sum_{\ell=1}^L D_{\text{JS}}\biggl(P^{(\ell)}(\hat{Z}|q_i) || P^{(\ell)}(\hat{Z}| q_j)\biggr)
\end{equation}


\noindent Concatenating $d$ for each language pair gives a distance matrix $\mathbf{D} \in \mathbb{R}^{n \times n}$ for $n$ languages. To this we can then apply UPGMA clustering to construct a dendrogram. UPGMA assumes 
equal root-to-tip distances, matching the time-calibrated property of the Bayesian phylogeny from \citet{heggarty2023language}, making dendrograms directly comparable.

\paragraph{LLMs, Data, and Distances}
To show results hold outside of a single model, our experiments look at 5 different LLM families:  BERT \citep{devlin_bert_2019, chung2020rethinking}, RoBERTa \citep{liu2019roberta}, mT5 \citep{xue2021mt5}, XLM \citep{conneau2020unsupervised, conneau2019cross}, and the Pythia models \citep{biderman_pythia_2023}, which make training checkpoints available. We focus on earlier LLMs in part because they come in explicitly monolingual and multilingual variants --- virtually all recent models are trained in a massively multilingual setting. Both BERT and RoBERTa come in English and multilingual variants, allowing us to look at how representations change as models are intentionally trained on broader sets of languages. Across all results and models, our estimates of conditional distributions $P(\hat{Z}| \text{language})$ are based on $7500$ sentences randomly sampled separately for each language from the multilingual C4 dataset - a broad crawl of internet data \citep{raffel2020exploring}. 


\subsection{Comparing LLM Representations with the Reference Tree}

We measure how well a model approximates similarity among languages by computing the distance between two hierarchical structures: the reference tree of language relations \citep{heggarty2023language}, and a dendrogram induced from LLM embeddings via density estimation and agglomerative clustering. We employ measures to capture three different aspects of hierarchical correspondence: the Cophenetic Correlation Coefficient, the Subtree Kernel, and the Path Kernel.

\paragraph{Rationale for Multiple Measures}
Each measure captures complementary aspects of the correspondence between representational and the similarity structure represented by the reference tree. The Cophenetic Correlation provides a global assessment of whether pairwise reference distances are preserved in the embedding space. The Path Kernel captures shared global hierarchical organization, testing whether the model encodes correct broad-to-narrow nesting of language subfamilies. The Subtree Kernel is the strictest test, identifying whether the model has exactly recovered local groupings,  including identical leaf sets and topology within each subfamily. Together, these measures characterize both overall fit and specific convergences/divergences between learned representations and the reference tree.

\paragraph{Cophenetic Correlation (CC)} 
measures the degree to which a hierarchical structure preserves pairwise distance relationships \citep{sokal1962comparison}. Given a hierarchical tree $\mathcal{T}$ over a set of items, the cophenetic distance $c_{ij}$ between items $i$ and $j$ is defined as the height of the most recent common ancestor (MRCA) at which $i$ and $j$ are first joined into a single cluster:
\begin{equation}
    c_{ij} = \text{height}(\text{MRCA}_{\mathcal{T}}(i,j))
\end{equation}
For trees with branch lengths, the cophenetic distance equals the sum of branch lengths from $i$ to their MRCA plus the sum from $j$ to their MRCA. For model-derived dendrograms from UPGMA, cophenetic distances correspond to the Jensen-Shannon divergences at which clusters were merged. We extract cophenetic distance matrices $\{c_{ij}^{\text{phylo}}\}$ and $\{c_{ij}^{\text{LLM}}\}$ from both the reference and model trees over the shared set of $n$ languages, then compute Pearson's correlation:

\begin{equation}
\rho_{\text{cophenetic}} = \text{Pearson}(\{c_{ij}^{\text{phylo}}\}_{i<j}, \{c_{ij}^{\text{LLM}}\}_{i<j})
\end{equation}

\noindent High CC indicates the model preserves relative proximity: language pairs closer in the phylogenetic tree remain proportionally closer in the model's representation space.

\paragraph{Subtree Kernel}
The subtree kernel quantifies tree similarity by counting shared 
subtree structures \citep{collins2001convolution, vishwanathan2010graph}. 
For two trees $\mathcal{T}_1$ and $\mathcal{T}_2$, we define an 
indicator function:
\begin{equation}
C(n_1, n_2) = \mathbbm{1} [\text{subtree}(n_1) \cong \text{subtree}(n_2)]
\end{equation}
where $\text{subtree}(n_1) \cong \text{subtree}(n_2)$ denotes 
subtree isomorphism: same topology and leaf set. For leaves, 
$C(n_1,n_2) = \mathbbm{1} [\text{label}(n_1) = \text{label}(n_2)]$. 
For internal nodes, $C(n_1,n_2) = 1$ if $\text{leaves}(n_1) = 
\text{leaves}(n_2)$ and all children match recursively, else 0. 
The kernel sums over all node pairs:
\begin{equation}
K_{\text{subtree}}(\mathcal{T}_1, \mathcal{T}_2) = 
    \sum_{n_1 \in \mathcal{T}_1} \sum_{n_2 \in \mathcal{T}_2} C(n_1, n_2)
\end{equation}
Normalized to be $\in [0,1]$. As with CC, higher scores indicate a model that better approximates the reference tree. High subtree kernel similarity indicates the model correctly identifies phylogenetic subfamilies (e.g., Romance, Germanic) as coherent clusters with matching internal structure, even if their global arrangement differs.

\paragraph{Path Kernel}
The path kernel measures tree similarity via shared root-to-leaf 
paths \citep{gartner2003survey, kashima2003marginalized}, capturing a global hierarchical organization. For a given language $\ell$, let $\text{path}_{\mathcal{T}}(\ell)$ denote the sequence of internal nodes from root to $\ell$. Path similarity is the normalized longest common prefix (LCP):
\begin{equation}
\text{sim}(\text{path}_{\mathcal{T}_1}(\ell), 
            \text{path}_{\mathcal{T}_2}(\ell)) = 
    \frac{|\text{LCP}(\text{path}_{\mathcal{T}_1}(\ell), 
                      \text{path}_{\mathcal{T}_2}(\ell))|}
         {\max(|\text{path}_{\mathcal{T}_1}(\ell)|, 
               |\text{path}_{\mathcal{T}_2}(\ell)|)}
\end{equation}
The kernel aggregates over all languages:
\begin{equation}
k_{\text{path}} = \frac{1}{|\mathcal{L}|}
                  \sum_{\ell \in \mathcal{L}} 
                  \text{sim}(\text{path}_{\mathcal{T}_1}(\ell), 
                              \text{path}_{\mathcal{T}_2}(\ell))
\end{equation}
yielding a similarity score in $[0,1]$ measuring global 
hierarchical alignment. High path kernel similarity indicates the model encodes correct hierarchical nesting: languages follow similar ancestral paths (e.g., French:  Indo-European → Romance → Italic), preserving multi-level family organization.

\section{Results \& Discussion}
The similarity structure for a single higher-performing model, XLM-Roberta, is provided in Figure 1 (Left) for visual comparison. Across our experiments, models broadly align representations with the reference tree based on \citet{heggarty2023language} (Figure 1: Right). 
XLM-Roberta's similarity structure does well in distinguishing Germanic, Romance, and Indo-Aryan language subfamilies from one another. The Slavic language family, too, would be clustered apart from other subfamilies were it not for the two Baltic languages incongruously dividing it, separating Slovenian from Serbian. Three independent languages — Greek, Welsh, and Albanian — pose a challenge to most models, with a recurring Greek-Welsh pairing in XLM-Roberta, m-BERT, and Pythia-160m that is particularly surprising given their lexical dissimilarity. The degree of alignment across each of our measures is provided along the x-axes in each facet of Figure 2. 

\paragraph{On Orthography} Tokenization in LLMs is sensitive to script, so differences in orthography likely contribute to certain patterns of similarity. For example, Slavic languages written in Cyrillic form a separate cluster from those that use Latin script. Yet results also demonstrate that script is not a determinant factor in LLM similarity representations. Hindi and Urdu — written in Devanagari and Perso-Arabic, respectively — are nonetheless grouped within the same subfamily across models, and Persian is consistently placed adjacent to North Indian languages despite its Arabic script. Moreover, exploratory analyses appropriately grouped Hebrew and Arabic, written in distinct scripts, together in a Semitic cluster, while neither Persian nor Urdu migrated toward this cluster despite sharing the Arabic script. This pattern suggests that orthography modulates similarity, but linguistic relatedness is the dominant signal LLMs latch on to. We therefore view tokenization as a property of the LLM whose interaction with similarity structure we are characterizing, rather than a confound exogenous to the model.

\paragraph{Similarity Structure Correlates with Performance}

To see what kinds of structure relate to performance, we look at representations from eight different LLMs correlating our three different distance measures with accuracy on a large-scale multilingual benchmark (shown in figure \ref{fig:xnli}). The Cophenetic Correlation (CC) and Path Kernel relate significantly to XNLI accuracy ($\rho=0.81, p=0.02$, $\rho=0.79, p=0.02$, respectively) while the subtree kernel does not ($\rho=0.383, p=0.349$). The Path Kernel and CC are more global measures, indicating whether the model has preserved the overall structure of the reference tree, capturing the higher-order categories like language subfamilies. By contrast, the subtree kernel looks at local structures, whether the exact neighbors in a model match. These results indicate that better-performing models more faithfully preserve linguistic similarity structure — as demonstrated here for Indo-European — and correctly organizing broader subfamily groupings matters more for performance than matching exact nearest-neighbour relationships.

\paragraph{English vs. multilingual Training Data}
BERT and RoBERTa were both trained predominantly, albeit not exclusively, on English data. Later variants, mBERT and XLM-RoBERTa, were explicitly trained as multilingual models using data from 104 and 100 languages, respectively. All four models have the same architecture and number of parameters. 
Figure \ref{fig:pretraining} shows how multilingual training data affects the similarity structure of models' representations. Both multilingual models achieve a higher match with our reference tree than their monolingual counterparts across all three measures.

\paragraph{Language-Level Structures Emerge Early in Training}
Figure \ref{fig:pretraining} looks at when the similarity structure across languages begins to emerge during pre-training by analysing checkpoints of two different sizes of Pythia models (160, and 410 million parameters. Both model sizes begin to align with the reference tree early in training, by step 100. This aligns with work on progressive differentiation, as in \citet[][]{saxe2019mathematical} whose account of semantic development shows how models learn broader distinctions first, before learning progressively subtler properties of the data. Compared to word or concept level distinctions, language identity is relatively broad, becoming clear early in training.


\section{Limitations and Future Work}

Future work needs to extend this analysis to other language families and benchmarks. As noted, systematic errors recur across models, including misplacement of independent languages and an unexplained Welsh-Greek pairing, warranting further investigation. The current analysis builds on similarity as a foundational construct in LLMs, earlier connectionist models, and psychology more generally. At the same time, the notion of similarity raises several well-known specters related to its context specificity \citep{medin1993respects, tversky1977}. Nonetheless, similarity is an invaluable metric for understanding the geometric representations of LLMs. The current results provide a case in point that extends its relevance to full languages. This suggests that historical changes that operate across most or all of a language may be analyzable as a general shift in similarity space, but this would require its own investigation.

\section{Conclusion}

The key contribution of the current work is two-fold: first, articulating the kind of higher-order similarity structure that emerges in LLM representations, as well as when and how it may arise. Second, showing a relationship between language-level similarity structures and a model's multilingual performance. This forms an extension of existing accounts of similarity's role in generalization at a far greater level of abstraction: similarity structure matters not just for generalizing at the level of words or sentences, but across entire languages. 

\printbibliography

\end{document}